\documentclass[conference]{IEEEtran}
\usepackage{times}
\usepackage{amssymb}

\usepackage[numbers]{natbib}
\usepackage{multicol}
\usepackage[bookmarks=true]{hyperref}

\usepackage{graphicx}
\usepackage{amsmath}
\graphicspath{{Figures}}
\DeclareGraphicsExtensions{.pdf,.png}
\usepackage{bigstrut}

\usepackage{csquotes}

\usepackage{algorithm, setspace}
\usepackage{algpseudocode}
\usepackage{url}
\usepackage{multirow}
\usepackage{float}
\usepackage{hyperref}
\usepackage{xcolor}

\begin{document}

\title{Shared Autonomy with Learned Latent Actions}

\author{\authorblockN{Hong Jun Jeon}
\authorblockA{Stanford University\\
hjjeon@stanford.edu}
\and
\authorblockN{Dylan P. Losey}
\authorblockA{Stanford University\\
dlosey@stanford.edu}
\and
\authorblockN{Dorsa Sadigh}
\authorblockA{Stanford University\\
dorsa@cs.stanford.edu}}

% avoiding spaces at the end of the author lines is not a problem with
% conference papers because we don't use \thanks or \IEEEmembership

% for over three affiliations, or if they all won't fit within the width
% of the page, use this alternative format:
% 
%\author{\authorblockN{Michael Shell\authorrefmark{1},
%Homer Simpson\authorrefmark{2},
%James Kirk\authorrefmark{3}, 
%Montgomery Scott\authorrefmark{3} and
%Eldon Tyrell\authorrefmark{4}}
%\authorblockA{\authorrefmark{1}School of Electrical and Computer Engineering\\
%Georgia Institute of Technology,
%Atlanta, Georgia 30332--0250\\ Email: mshell@ece.gatech.edu}
%\authorblockA{\authorrefmark{2}Twentieth Century Fox, Springfield, USA\\
%Email: homer@thesimpsons.com}
%\authorblockA{\authorrefmark{3}Starfleet Academy, San Francisco, California 96678-2391\\
%Telephone: (800) 555--1212, Fax: (888) 555--1212}
%\authorblockA{\authorrefmark{4}Tyrell Inc., 123 Replicant Street, Los Angeles, California 90210--4321}}

\maketitle

\begin{abstract}
Assistive robots enable people with disabilities to conduct everyday tasks on their own. However, these tasks can be complex, containing both coarse reaching motions and fine-grained manipulation. For example, when eating, not only does one need to move to the correct food item, but they must also precisely manipulate the food in different ways (e.g., cutting, stabbing, scooping). Shared autonomy methods make robot teleoperation safer and more precise by arbitrating user inputs with robot controls. However, these works have focused mainly on the high-level task of \emph{reaching} a goal from a discrete set, while largely ignoring \emph{manipulation} of objects at that goal.
Meanwhile, dimensionality reduction techniques for teleoperation map useful high-dimensional robot actions into an intuitive low-dimensional controller, but it is unclear if these methods can achieve the requisite \emph{precision} for tasks like eating.
Our insight is that---by combining intuitive embeddings from learned latent actions with robotic assistance from shared autonomy---we can enable \emph{precise assistive manipulation}. In this work, we adopt learned latent actions for shared autonomy by proposing a new model structure that changes the meaning of the human's input based on the robot's confidence of the goal. We show convergence bounds on the robot's distance to the most likely goal, and develop a training procedure to learn a controller that is able to move between goals even in the presence of shared autonomy. We evaluate our method in simulations and an eating user study. See videos of our experiments here: \href{https://youtu.be/7BouKojzVyk}{\color{orange}{https://youtu.be/7BouKojzVyk}}.

\end{abstract}

\IEEEpeerreviewmaketitle

\section{Introduction}

There are nearly one million American adults living with physical disabilities that need external assistance when eating \cite{taylor2018americans}. Physically assistive robots---such as wheelchair-mounted robotic arms---promise to help these people eat independently, without relying on caregivers \cite{jacobsson2000people, mitzner2018closing}. We envision a future where users teleoperate assistive robots (e.g., through joysticks \cite{herlant2016assistive}, sip-and-puffs \cite{argall2018autonomy}, or brain-computer interfaces \cite{muelling2017autonomy}) to seamlessly perform complex and dexterous eating tasks. 

\begin{figure}[t]
	\begin{center}
		\includegraphics[width=0.8\columnwidth]{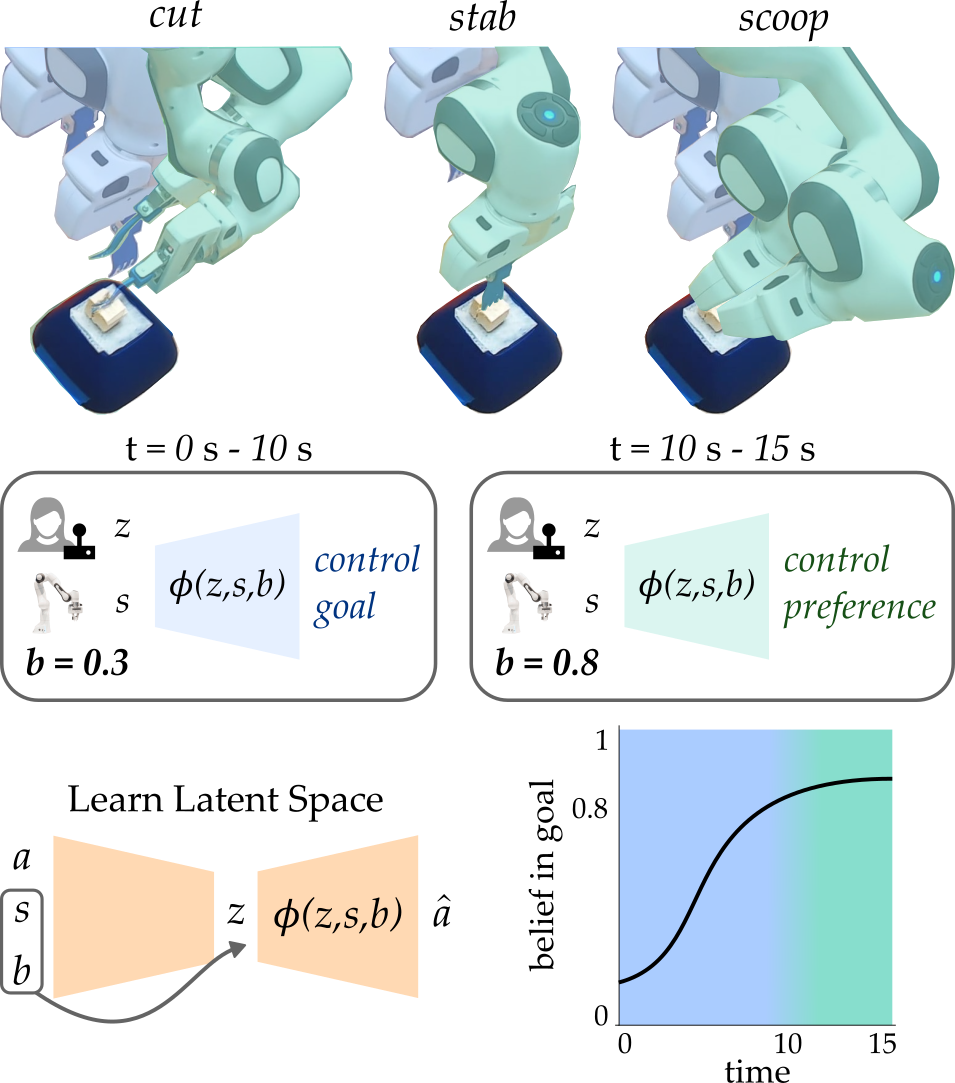}
	 	\vspace{-0.5em}		
		\caption{User teleoperating an assistive robot to perform eating tasks. The human starts by controlling the robot's high-level motion. As the robot gets more confident about the human's goal, the meaning of the inputs becomes more refined, and the human precisely adjusts the robot's movement.}
		\label{fig:front}
	\end{center}
	\vspace{-2em}
\end{figure}

For instance, imagine that you are controlling an assistive robotic arm to get a bite of tofu. You have a high-level goal: you want to guide the robot to reach for the tofu on the table in front of you. But just reaching the tofu is not enough; once there, you also need to \textit{precisely} manipulate the arm to cut off a piece and pick it up with your fork (see Fig.~\ref{fig:front}). An effective robotic partner should assist with both the high-level reaching motions and the fine-grained manipulation tasks.

We will refer to the human's high-level objectives as \textit{goals} (e.g., reaching the tofu), and their low-level manipulation as \textit{preferences} (e.g., cutting, stabbing, or scooping). Completing a task according to your goals and preferences is challenging---particularly because today's assistive robots are teleoperated using \textit{low-dimensional} interfaces \cite{herlant2016assistive, argall2018autonomy, muelling2017autonomy}, while these tasks require high-dimensional, coordinated, and precise control. 

Shared autonomy can make eating tasks easier by predicting the human's intent and then augmenting their input \cite{javdani2018shared, dragan2013policy, jain2019probabilistic}. But current works focus on \textit{goals}---i.e., helping the human move the robot between a set of discrete options---and do not provide assistance \textit{after} reaching the goal, when the human must control the robot along a continuum of preferences.

Another approach is to develop better interfaces for directly controlling the robot. For example, the robot can use demonstrations to \textit{learn} an intelligent mapping between the human's low-dimensional inputs and the robot's high-dimensional actions \cite{losey2019controlling}. This approach makes sense when moving along the continuum of \textit{preferences}; but the robot does not provide any additional guidance, so that any \textit{imperfect or noisy} human inputs will move the robot away from the precise goal.

Viewed together, humans must be able to intuitively control the robot while performing complex and precise manipulation tasks. These tasks involve moving between discrete, high-level goals and fine-tuning along continuous, low-level preferences.
\begin{center}
    \textit{We combine learning intuitive embeddings with shared autonomy to enable precise assistive manipulation.}
\end{center}
We view these methods as complementary types of assistance. Shared autonomy \textit{constrains} the robot to high-level goals, while intuitive mappings \textit{embed} the robot's motion into a sub-manifold of preferences. Returning to our eating example: the human starts by guiding the robot arm towards the tofu. As the robot becomes more confident about this goal, shared autonomy completes the motion, and the user's inputs transition to control the robot's fine-grained preferences (see. Fig.~\ref{fig:front}). 

Overall, we make the following contributions:

\noindent\textbf{Controlling Goals and Preferences.} We formalize assistive tasks with goals and preferences and introduce the properties that intuitive interfaces should have during these tasks. Next, we propose a new model structure where the meaning of the human's inputs changes based on the robot's confidence.

\noindent\textbf{Balancing Convergence with Change.} We theoretically identify a convergence bound on the robot's distance from the most likely goal. To ensure that users are not trapped at the wrong goal, we add a novel entropy term that encourages versatility, and test this with a spectrum of simulated humans.

\noindent\textbf{Conducting User Study with Eating Task.} Participants teleoperate a 7 degree-of-freedom (\textit{DoF}) robot arm with a 2-DoF joystick. We compare our approach to state-of-the-art baselines, and measure precision with and without both shared autonomy and the intuitive embedding.
\section{Related Work}

We combine latent actions---a representation learning technique for discovering low dimensional input spaces for controlling a high dimensional system---with shared autonomy, a well established paradigm for incorporating both human and robot inputs to control a system.

\smallskip
\noindent\textbf{Application Area -- Assistive Robotics.} Among the challenges of eating is the ability to prepare and transport food \cite{jacobsson2000people}, a skill that assistive robotics has tried to enable. Researchers have developed robot policies that can autonomously manipulate and deliver different types of food to users with disabilities \cite{feng2019robot, park2019toward}. However, designing a fully autonomous system to handle a task as variable and personalized as eating is exceedingly challenging: one size \textit{does not} fit all. We therefore develop a \emph{partially} autonomous algorithm that allows users to perform precise manipulation tasks with a robotic arm.

\smallskip
\noindent\textbf{Shared Autonomy.} Our partially autonomous system falls under shared autonomy, a framework in which the robot receives human user inputs and combines them appropriate autonomous inputs for a safer and more efficient outcome.
Many shared autonomy algorithms provide robotic assistance to users who must reach for a goal objects in the environment \cite{dragan2013policy, javdani2018shared, muelling2017autonomy, gopinath2016human, 10.1145/3171221.3171287, huang2016anticipatory}. Several of these works infer human intent by maintaining a belief over the set of possible goals, and using human inputs as evidence in a Bayesian framework to continually update this belief \cite{dragan2013policy, javdani2018shared, muelling2017autonomy,sadigh2016information,sadigh2018planning}. The algorithms often apply more assistance as confidence in a goal increases.

Other works in shared autonomy propose or learn suitable dynamics models to translate user inputs to robot actions \cite{rakita2017motion, reddy2018shared, DBLP:journals/corr/abs-1805-08010}. The learned latent actions in our approach mirror the work of Reddy \textit{et. al} \cite{reddy2018shared}, but have the added challenge of mapping a \emph{low} dimensional input to a \emph{high} dimensional action.

\smallskip
\noindent\textbf{Latent Actions.} Controlling robots is difficult when they contains many degrees of freedom (DoFs). To control the entire system, joysticks often include a button to toggle between different modes, each of which controls a subset of the full system. This can be extremely taxing for the users \cite{herlant2016assistive}. Prior work has tried to prune away unnecessary control axes through Principal Components Analysis (PCA) \cite{ciocarlie2009hand, 5401049}. More recently, researchers have learned \emph{non-linear} mappings to control high dimensional robot arms through low dimensional inputs \cite{losey2019controlling}. There is also extensive work in learning latent representations for components in RL \cite{pmlr-v97-chandak19a, DBLP:journals/corr/abs-1806-02813, DBLP:journals/corr/WatterSBR15, DBLP:journals/corr/abs-1903-01973, DBLP:journals/corr/abs-1805-07914}. These methods use autoencoders \cite{kingma2013auto, doersch2016tutorial} to learn mappings between structurally sparse high dimensional data and low dimensional embedding spaces without supervision. We build on work in learned latent actions \cite{losey2019controlling} and task-conditioning \cite{noseworthytask} to learn an embedding for teleoperation \emph{in the presence} of shared autonomy.

\section{Using Latent Actions \& Shared Autonomy}

We explore tasks where the robot's movements naturally become more refined and precise over time. Recall our eating example: users start by indicating where the robot should go (e.g., \textit{reach for the tofu}), and then control what the robot does at that goal (e.g., \textit{cut off a piece and pick it up with the fork}). Completing these tasks is particularly challenging with assistive robots, where the user must interact with a low-dimensional control interface (e.g., a joystick).

\smallskip

\noindent \textbf{Overview.} In this section, we propose an algorithm that \textit{refines} the robot's assistance. At the start of the task, the human's low-DoF inputs coarsely move the robot towards a high-level \textit{goal} (e.g., reaching the tofu). Once the robot reaches this goal, the low-DoF inputs change meaning to precisely manipulate along the human's low-level \textit{preferences} (e.g., cutting a piece). We leverage latent actions to learn this changing mapping: specifically, we learn a decoder $\phi(\cdot)$ that enables the human to control a spectrum of goal-directed motions and fine-grained preferences. In order to guide the robot to the user's goal---and maintain this goal while the human focuses on conveying their preference---we apply shared autonomy.

\smallskip

\noindent\textbf{Formulation.} We formulate the human's task as a Markov Decision Process (MDP) $\mathcal{M} = \langle \mathcal{S}, \mathcal{A}, \mathcal{T}, \mathcal{G}, \Theta, R, \gamma \rangle$. Let $s \in \mathcal{S} \subseteq \mathbb{R}^n$ be the robot's state and let $a \in \mathcal{A} \subseteq \mathbb{R}^m$ be the robot's action: when the robot takes action $a$ in state $s$, it transitions according to $\mathcal{T}(s, a)$. The human has a high-level goal $g^* \in \mathcal{G}$ and low-level preference $\theta^* \in \Theta$. Together, the goal (e.g., \textit{reaching the tofu}) and preference (e.g., \textit{cutting the tofu}) determine the robot's reward function: the robot should maximize $R(s, g^*, \theta^*)$ with the discount factor $\gamma \in [0, 1)$. 

The space of candidate goals $\mathcal{G}$ is discrete and \textit{known} by the robot \textit{a priori}. We let $b \in \mathcal{B}$ denote the robot's belief over this space of candidate goals, where $b(g) = 1$ indicates that the robot is convinced that $g$ is the human's desired goal.

The space of preferences $\Theta$ is continuous and \textit{unknown} by the robot. We do not maintain a belief here; instead, we assume the robot has access to $\mathcal{D}$, a dataset of relevant demonstrations (e.g., examples of reaching for and then manipulating the tofu). These demonstrations consist of state-action-belief tuples: $\mathcal{D} = \{(s^0, a^0, b^0), (s^1, a^1, b^1), \ldots\}$.

\smallskip

\noindent\textbf{Dynamics.} The human teleoperates the robot by using a low-dimensional interface (e.g., a joystick). Let $z \in \mathcal{Z} \subset \mathbb{R}^d$ be the human's input---where $d < m$---and let $\phi(\cdot)$ be a \textit{decoder} function that maps these low-dimensional human inputs into the robot's action space, $\mathcal{A}$. We denote the resulting action as $a_h \in \mathcal{A}$. The robot's overall behavior combines this input action and $a_r \in \mathcal{A}$, the robot's assistive guidance \cite{dragan2013policy, newman2018harmonic}:
\begin{equation} \label{eq:M1}
    a = (1 - \alpha) \cdot a_h + \alpha \cdot a_r
\end{equation}
$\alpha \in [0, 1]$ parameterizes the trade-off between direct teleoperation ($\alpha = 0$) and complete autonomy ($\alpha = 1$).

\smallskip

\noindent\textbf{Problem Statement.} Our objective is to intuitively \textit{decode} the human's low-DoF inputs by learning $\phi(\cdot)$, and then combine these inputs with the robot's \textit{autonomous} high-DoF actions $a_r$. This problem is made challenging by the need to assist users as they control dexterous robots along coarse, goal-directed movement and precise, preferred manipulations. 

\subsection{Learned Latent Actions (LA)} \label{sec:LA}

Latent actions refer to low-dimensional representations of high-dimensional actions that are learned through dimensionality reduction techniques \cite{losey2019controlling, DBLP:journals/corr/abs-1903-01973, jonschkowski2014state}. Given a set of demonstrated motions, the robot embeds the high-DoF actions into a latent action space, and then decodes these latent actions to reconstruct the original action (see Fig.~\ref{fig:front}, bottom). Previous works have leveraged latent actions for intuitive low-DoF control of assistive robots, where latent actions enable users to express their desired high-DoF motion \cite{losey2019controlling, ciocarlie2009hand, 5401049}.

Unlike these prior works, we recognize that often the \textit{meaning} of latent actions changes within a precision task. Imagine that you are using a 1-DoF interface to perform the task in our eating example. At the start of the task, you need pressing left and right on the joystick (i.e., $z$) to move the robot towards the tofu. But as the robot approaches the tofu, you no longer need to keep moving towards a goal; instead, you need to use those same joystick inputs to carefully align the orientation of the fork, so that you can cut off a piece.

While latent actions provide an intuitive way to convey the user's intent, there are often more actions to convey than the latent space can embed. We thus need latent actions that can \textit{convey different types of meanings}---indicating the human's goal, preference, or some combination of the two.

\smallskip

\noindent\textbf{Conditioning on Belief.} In order to learn latent action spaces that can continuously alternate between controlling high-level goals and fine-grained preferences, we will condition on the robot's current \textit{context}. This context includes the state $s$ -- the configuration that the robot is in -- as well as the \textit{belief} $b$ -- the robot's confidence in the goal. Intuitively, conditioning on belief enables the \textit{meaning} of latent actions to change based on the robot's confidence. As a result of this proposed structure, latent actions \textit{purely indicate the desired goal} when the robot is unsure; and once the robot is confident about the human's goal, latent actions change to \textit{convey the preferred manipulation}. We design and enforce models that decode the meaning of latent actions based on context: $\phi : \mathcal{Z} \times \mathcal{S} \times \mathcal{B} \rightarrow \mathcal{A}$.

\smallskip

\noindent\textbf{Reconstructing Actions.} We now have a robot that decodes the user's input based on its state and confidence; but how do we ensure the decoded actions are \textit{accurate}? Put another way, how do we ensure that the robot learns latent actions that can actually move the high-dimensional robot towards the human's goal and then correctly manipulate the object? To resolve this problem, we turn to the dataset $\mathcal{D}$, which contains examples of these desired, high-DoF actions. Specifically, consider the state-action-belief tuple $(s, a, b) \in \mathcal{D}$: we want to learn a latent action space $\mathcal{Z}$ such that given $s$, $b$, and some $z \in \mathcal{Z}$, the robot reconstructs the demonstrated action $a$. Let $a_h \in \mathcal{A}$ be the reconstructed action, where $a_h = \phi(z,s,b)$, and let $e_a = a_h - a$ be the error between the reconstructed and demonstrated actions. To learn latent spaces that accurately reconstruct the demonstrated actions, we explore models that actively attempt to minimize the reconstruction error $\|e_a\|^2$ in the loss function.

\subsection{Latent Actions with Shared Autonomy (LA + SA)}

Latent actions provide an expressive mapping between low-dimensional user inputs and high dimensional robot actions. But controlling a robot with latent actions alone still presents a challenge: any \textit{imprecision} or \textit{noise} in either the user's inputs or latent space is reflected in the decoded actions. Recall our eating example: when the user is trying to guide the robot towards their preferred cutting motion, their inputs should not unintentionally cause the robot arm to drift away from the tofu or suddenly jerk into the table. Accordingly, we here leverage shared autonomy to facilitate \textit{precise} robot motions, which assist the human towards their goals, and then maintain these goals as the human focuses on their preferences.

\smallskip

\noindent \textbf{Providing Assistance.} Recall that the robot applies assistance via action $a_r$ in Eq. (\ref{eq:M1}). In order to assist the human, the robot needs to understand the human's \textit{intent}---i.e., which goal they want to reach. The robot's understanding of the human's intended goal is captured by the belief $b$, and we can leverage this belief to select an assistive action $a_r$. Similar to \cite{dragan2013policy} and \cite{javdani2018shared}, let the robot provide assistance towards each discrete goal $g \in \mathcal{G}$ in proportion to the robot's confidence in that goal\footnote{Our approach is not tied on this particular instantiation of shared autonomy. Other instances of shared autonomy can similarly be used.}:
\begin{equation} \label{eq:M2}
    a_r = \sum_{g \in \mathcal{G}} b(g) \cdot (g - s)
\end{equation}
So now---if the robot has a uniform prior over which morsel of food the human wants to eat---$a_r$ guides the robot to the center of these morsels. And---when the human indicates a desired morsel---$a_r$ guides the robot towards that target.

\smallskip

\begin{algorithm}[t]
\caption{Latent Actions \& Shared  Autonomy (\textit{LA+SA})}
  \label{alg:LA+SA}

\begin{algorithmic}[1] \setstretch{1.0}
    \State Given a discrete set of goals $\mathcal{G}$ and a dataset of example trajectories $\mathcal{D} = \{(s^0,a^0,b^0), (s^1,a^1,b^1), \ldots \}$
    \State Train a model on $\mathcal{D}$ to learn the decoder $\phi(z,s,b)$
    \For{$t \gets 1, 2, \ldots, T$}
        \State Set $z^t$ as the human's low-DoF input
        \State $a_h^t \gets \phi(z^t, s^t, b^t)$ \Comment{map $z^t$ to high-DoF action}
        \State $a_r^t \gets \sum_{g \in \mathcal{G}} b^t(g) \cdot (g - s^t)$  \Comment{get robot assistance}
        \State $a^t \gets (1 - \alpha) \cdot a_h^t + \alpha \cdot a_r^t$ \Comment{blend both $a_h$ and $a_r$}
        \State $b^{t+1} \propto P(a_h | s, g)P(g)$ \Comment{update belief over goals}
        \State $s^{t+1} \sim \mathcal{T}(s^t, a^t)$ \Comment{take action and transition states}
    \EndFor
\end{algorithmic}
\end{algorithm}

\noindent\textbf{Algorithm.} Our approach for combining latent actions (\textit{LA}) with shared autonomy (\textit{SA}) is summarized in Algorithm~\ref{alg:LA+SA}. We emphasize that \textit{LA+SA} is different from either latent actions or shared autonomy alone. Without latent actions, a robot using shared autonomy must rely on predetermined, one-size-fits-all mappings from $z$ to $a_h$. Without shared autonomy, latent actions require perfect teleoperation to reach and maintain goals. Put another way: shared autonomy constrains the robot to goals, while latent actions embed the robot's motions into a continuous sub-manifold of preferences.

\section{Theoretical Analysis}

We propose \textit{LA+SA} as an approach for tasks involving goals and preferences. Both latent actions and shared autonomy have an \textit{independent} role within this method: but how can we be sure that the \textit{combination} of these tools will remain effective? Returning to our eating example---if the human inputs latent actions, will shared autonomy correctly guide the robot to the desired morsel of food? What if the human has multiple goals in mind (e.g., getting a chip and then dipping it in salsa)---can the latent actions change goals even when shared autonomy is confident? And what if the environment changes---how do we transfer the learned latent actions to new goals?

\subsection{Converging to the Desired Goal} \label{sec:convergence}

We first explore how \textit{LA+SA} ensures that the human reaches their desired goal. Consider the Lyapunov function:
\begin{equation} \label{eq:T1}
    V(t) = \frac{1}{2} \|e(t)\|^2, \quad e(t) = g^* - s(t)
\end{equation}
where $e$ denotes the error between the robot's current state $s$ and the human's goal $g^*$. We want the robot to choose actions that minimize Eq.~(\ref{eq:T1}) across a spectrum of user skill levels and teleoperation strategies. Let us focus on the common setting in which $s$ is the robot's joint position and $a$ is the joint velocity, so that $\dot{s}(t) = a(t)$. Taking the derivative of Eq.~(\ref{eq:T1}) and substituting in this transition function, we reach\footnote{For notational simplicity we choose $\alpha = 0.5$, so that both human and robot inputs are equally weighted. Our results generalize to other $\alpha$.}:
\begin{equation} \label{eq:T2}
    \dot{V}(t) = -\frac{1}{2}e^\top \Big[\phi(z,s,b) + \sum_{g \in \mathcal{G}}b(g)\cdot (g - s)\Big]
\end{equation}
We want Eq.~(\ref{eq:T2}) to be negative, so that $V$ (and thus the error $e$) decrease over time. A sufficient condition for $\dot{V} < 0$ is:
\begin{equation} \label{eq:T3}
    b(g^*) \cdot \|e\| > \|\phi(z,s,b)\| + \sum_{g \in \mathcal{G}'} b(g)\cdot \|g - s\|
\end{equation}
where $\mathcal{G}'$ is the set of all goals except $g^*$. As a final step, we bound the magnitude of the decoded action, such that $\| \phi(\cdot)\| < \sigma_h$, and we define $\sigma_r$ as the distance between $s$ and the furthest goal: $\sigma_r = \max_{g \in \mathcal{G}'} \|g - s\|$. Now we have $\dot{V} < 0$ if:
\begin{equation} \label{eq:T4}
    b(g^*)\cdot \|e\| > \sigma_h + \big(1 - b(g^*)\big)\cdot \sigma_r
\end{equation}
We define $\delta := \sigma_h + \big(1 - b(g^*)\big)\cdot \sigma_r$. We therefore conclude that \textit{LA+SA} yields \textit{uniformly ultimately bounded stability} about the human's goal, where $\delta$ affects the \textit{radius} of this bound \cite{spong2006robot}. As the robot's confidence in $g^*$ increases, $\delta \rightarrow \sigma_h$, and the robot's error $e$ decreases so long as $\|e(t) \| > \sigma_h$. Intuitively, \textit{LA+SA} guarantees that the robot will move to some ball around the human's goal $g^*$, and the radius of that ball decreases as the robot becomes more confident.

\subsection{Changing Goals} \label{sec:entropy}

Our analysis in Sec.~\ref{sec:convergence} suggests that the robot becomes \textit{constrained} to a region about the most likely goal. This works well when the human correctly conveys their intentions to the robot---but what if the human makes a mistake, or changes their mind? How do we ensure that the robot is not \textit{trapped} at an undesired goal? Re-examining Eq.~(\ref{eq:T4}), it is key that---at every $(s,b)$ pair---the human can convey sufficiently large actions $\|\phi(z,s,b)\|$ towards their preferred goal, ensuring that $\sigma_h$ does not decrease to zero. Put another way, the human must be able to \textit{increase} the radius of the bounding ball, reducing the constraint imposed by shared autonomy.

To encourage the robot to learn latent actions that increase this radius, we introduce an additional term into our model's loss function. We reward the robot for learning latent actions that have high \textit{entropy} with respect to the goals; i.e., in a given context $(s,b)$ there exist latent actions $z$ that cause the robot to move towards \textit{each} of the goals $g \in \mathcal{G}$. Define $p_{(s, b)}(g)$ as proportional to the total \emph{score} $\eta$ accumulated for goal $g$:
\begin{equation} \label{eq:T5}
    p_{(s, b)}(g) \propto \sum_{z\in \mathcal{Z}} \eta(g, s, b, z)
\end{equation}
where the score function $\eta$ indicates how well action $z$ taken from state $(s, b)$ conveys the intent of moving to goal $g$, and the distribution $p_{(s,b)}$ over $\mathcal{G}$ captures the proportion of latent actions $z$ at state $(s, b)$ that move the robot toward each goal. Intuitively, $p_{(s,b)}$ captures the comparative ease of moving toward each goal: when $p_{(s,b)}(g) \rightarrow 1$, the human can easily move towards goal $g$ since \emph{all} latent actions at $(s,b)$ induce movement towards goal $g$ and consequently, \emph{no} latent actions guide the robot towards any other goals. We seek to \textit{avoid} learning latent actions where $p_{(s,b)}(g) \rightarrow 1$, because in these scenarios the teleoperator \textit{cannot} correct their mistakes or move towards a different goal! Recall from Sec.~\ref{sec:LA} that the model should minimize the reconstruction error, $e_a = a_h - a$. We now argue that the model should additionally maximize the Shannon entropy of $p$, so that the loss function becomes:
\begin{equation} \label{eq:T6}
   \mathcal{L} = \|e_a\|^2 + \lambda \cdot \sum_{g \in \mathcal{G}} p(g) \log{p(g)} 
\end{equation}
Here the hyperparameter $\lambda > 0$ determines the relative trade-off between reconstruction error and latent action entropy. For clarity, we emphasize that this loss function $\mathcal{L}$ is leveraged offline, when training a model to learn the decoder $\phi(\cdot)$.

\subsection{Introducing New Goals} \label{sec:invariance}

We have covered how the robot can reach and change goals during the task---but what about situations where new goals are introduced \textit{dynamically}? For instance, imagine that in our eating scenario a new plate of food is set in front of the user. The decoder $\phi(\cdot)$ has already been trained using the dataset $\mathcal{D}$, which does not include example trajectories reaching for this plate. Accordingly, the latent action space may not contain actions that move the robot towards this new plate, preventing the human from interacting with this new goal!

We resolve this issue by leveraging the goals that the robot has already seen, \textit{without collecting} new demonstrations \textit{or retraining} the latent space. Let $g$ be the new goal, and define $h(s,b) \rightarrow (\hat{s},\hat{b})$ as a function that maps the robot's current context $(s,b)$ to an \textit{equivalent} context with respect to the previously seen goal $\hat{g}$. As a simple example, $h$ could project the robot's current state $s$ to a straight-line path between the start and $g$, and output the equivalent state $\hat{s}$ along the straight-line path between the start and $\hat{g}$ (while assigning the same confidence to $\hat{g}$ as the robot currently has over $g$). Using $h$, the overall process is as follows: (a) Convert to an equivalent context $(\hat{s},\hat{b})$ where training data exists. (b) Decode the user's latent input in this equivalent context to identify the high-DoF action $\hat{a}_h = \phi(z,\hat{s},\hat{b})$. (c) Transform $\hat{a}_h$ back to the original state to obtain the commanded action $a_h$. Robots can harness the models they have already learned with newly added goals, so that---if the robot has learned to pour, scoop, and stir at one bowl---the human still has these same latent actions available at a second bowl that has just been introduced.
\section{Simulations}

Our theoretical analysis highlights the benefits of combining latent actions with shared autonomy. However, it is not clear how this approach will work when interacting with a \textit{spectrum of different users}. In this section we test our algorithm in a controlled environment with simulated humans. We compare models for learning latent actions with and without shared autonomy, and we simulate teleoperators with various levels of expertise and learning rates. We will investigate if \textit{different types of users} can interact with our algorithm to \textit{precisely} reach and change goals consistent with their preferred trajectory.

\smallskip

\noindent\textbf{Model.} We test latent actions used by themselves (\textit{LA}), as well as latent actions combined with shared autonomy (\textit{LA+SA}). For both \textit{LA} and \textit{LA+SA} we learn the latent space with an autoencoder conditioned on state and belief (as described in Sec.~\ref{sec:LA}). Building on our analysis from Sec.~\ref{sec:entropy}, we also test \textit{LA+SA+Entropy}, where the autoencoder leverages Eq.~(\ref{eq:T6}) to additionally reward entropy in the learned latent space.

\smallskip

\begin{figure}[t]
	\begin{center}
		\includegraphics[width=1\columnwidth]{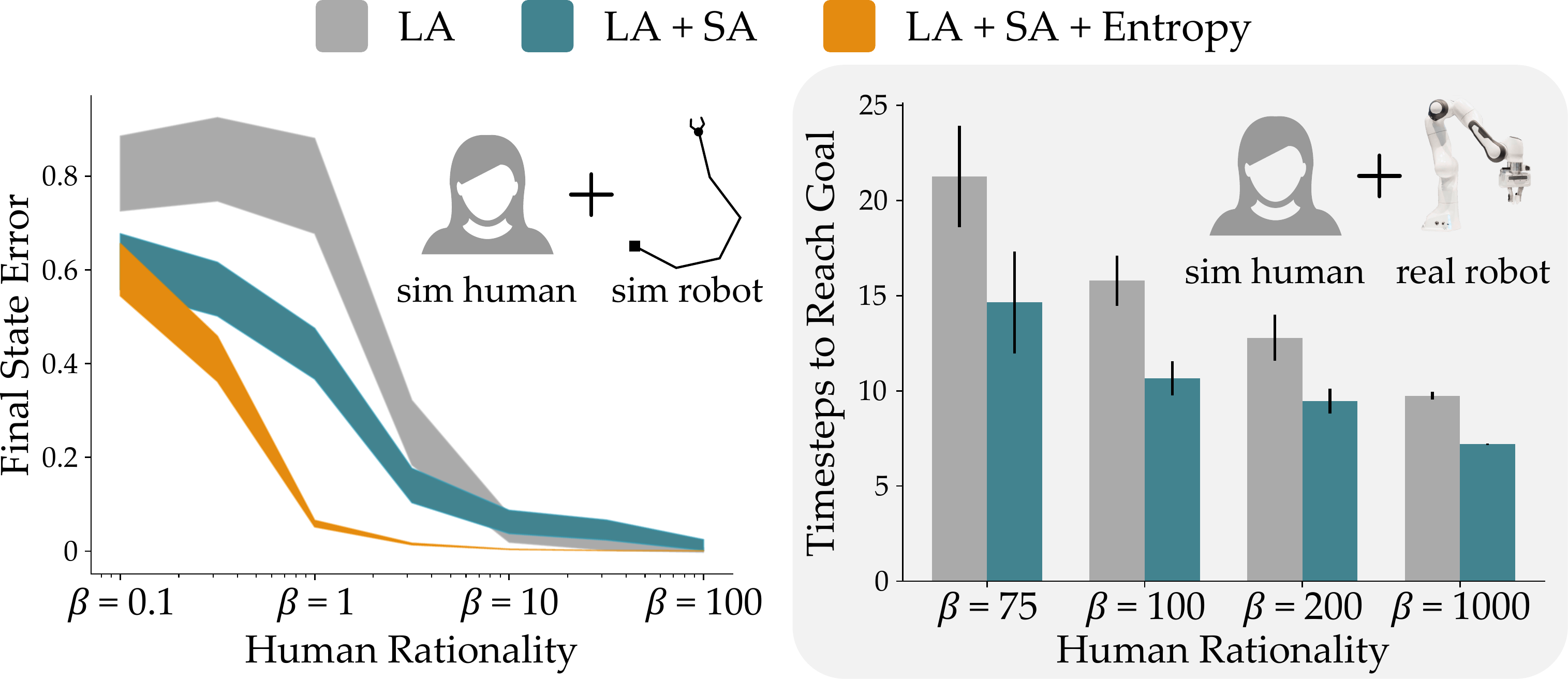}
	 	\vspace{-1.5em}
		\caption{Simulated humans for different levels of rationality. As ${\beta \rightarrow \infty}$, the human's choices approach optimal inputs. \textit{Final State Error} (in all plots) is normalized by the distance between goals. Introducing shared autonomy (\textit{SA}) improves the convergence of latent actions (\textit{LA}), particularly when the human teleoperator is noisy and imperfect.}
		\label{fig:sim1}
	\end{center}
	\vspace{-2em}
\end{figure}

\begin{figure*}[t]
	\begin{center}
		\includegraphics[width=1.8\columnwidth]{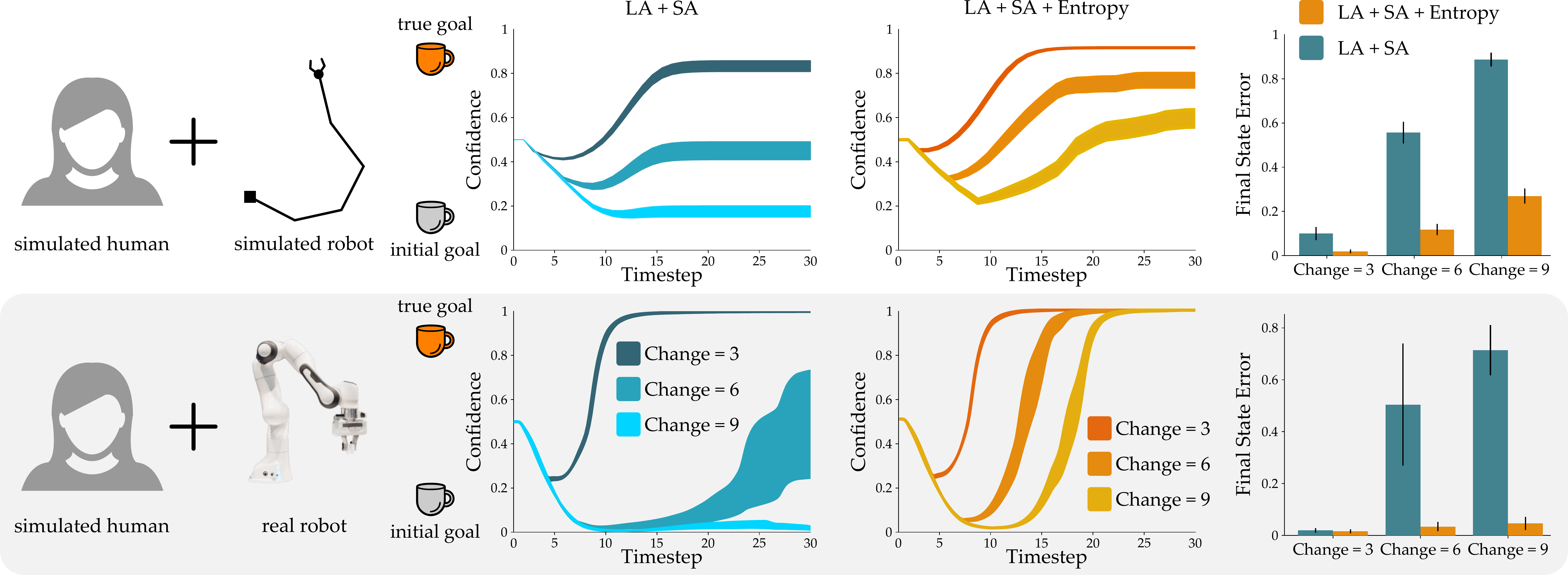}
		\vspace{-0.5em}
		\caption{Simulated humans that change their intended goal part-way through the task. \textit{Change} is the timestep where this change occurs, and \textit{Confidence} refers to the robot's belief in the human's true goal. Because of the constraints imposed by shared autonomy, users need latent actions that can overcome misguided assistance and move towards a less likely (but correct) goal. Encouraging entropy in the learned latent space (\textit{LA+SA+Entropy}) enables users to switch goals.}
		\label{fig:sim2}
	\end{center}
	\vspace{-2em}
\end{figure*}

\noindent\textbf{Environments.} We implement these models on both a \textit{simulated} and a \textit{real} robot. The simulated robot is a $5$-DoF planar arm, and the real robot is a $7$-DoF FrankaEmika. For both robots, the state $s$ captures the current joint position, and the action $a$ is a change in joint position, so that: $s^{t+1} = s^t + \Delta \cdot a^t$.

\smallskip

\noindent\textbf{Task.} We consider a manipulation task where there are two coffee cups in front of a robot arm. The human may want to reach either cup (i.e., goals), and grasp that cup along a continuous spectrum from the top to the side (i.e., preferences). We embed the robot's high-DoF actions into a $1$-DoF input space: the simulated users had to convey both their goal and preference \textit{only by pressing left and right} on the joystick.

\smallskip

\noindent\textbf{Simulated Humans.} The users attempting to complete this task are \textit{approximately optimal}, and make decisions that guide the robot accordingly to their goal $g^*$ and preference $\theta^*$. Let $s^*$ be the final pose that the human wants the robot to reach: $s^*$ is based on both the position of their desired coffee cup ($g^*$) and the orientation of their preferred grasp ($\theta^*$). The humans have reward function $R = -\|s^* - s\|^2$, and choose latent actions $z$ to move the robot directly towards $s^*$:
\begin{equation} \label{eq:S1}
	p(z) \propto \exp{ \Big\{ - \beta(t) \cdot \| s^* - (s + \phi(z, s, b)) \|^2 \Big\}}
\end{equation}
Within Eq.~(\ref{eq:S1}), $\beta \geq 0$ is a temperature constant that affects the user's \textit{rationality}. When $\beta \rightarrow 0$, humans select increasingly random $z$, and when $\beta \rightarrow \infty$, humans always choose the $z$ that moves the robot arm along their goal and preference. We simulate different types of users by varying $\beta(t)$.

\subsection{Users with Fixed Expertise}

We first simulate humans that have \textit{fixed} levels of expertise. Here expertise is captured by $\beta$ from Eq.~(\ref{eq:S1}): users with high $\beta$ are proficient, and rarely make mistakes with noisy inputs. We anticipate that all algorithms will perform similarly when humans are always perfect or completely random---but we are particularly interested in the spectrum of users \textit{between} these extremes, who frequently \textit{mis-control} the robot.

Our results relating $\beta$ to performance are shown in Fig.~\ref{fig:sim1}. In accordance with our convergence result from Sec.~\ref{sec:convergence}, we find that introducing shared autonomy helps humans reach their desired grasp more quickly, and with less final state error. The performance difference between \textit{LA} and \textit{LA+SA} decreases as the human's expertise increases---looking specifically at the real robot simulations, \textit{LA} takes $45\%$ more time to complete the task than \textit{LA+SA} at $\beta = 75$, but only $30\%$ more time when $\beta = 1000$. We conclude that shared autonomy improves performance across all levels of expertise, both when latent actions are trained with and without \textit{Entropy}.

\subsection{Users that Change their Mind}

One downside of shared autonomy is over-assistance: like we discussed in Sec.~\ref{sec:entropy}, the robot may become \textit{constrained} at likely (but incorrect) goals. To examine this adverse scenario we simulate humans that \textit{change} which coffee cup they want to grasp after $N$ timesteps. These simulated users intentionally move towards the \textit{wrong} cup while $t \leq N$, and then try to reach the correct cup for the rest of the task.
We model humans as near-optimal immediately after changing their mind about the goal, following Eq.~(\ref{eq:S1}) with a high $\beta$ value. 

We visualize our results in Fig.~\ref{fig:sim2}. When the latent action space is trained only to minimize reconstruction loss (\textit{LA+SA}), users cannot escape the shared autonomy constraint around the wrong goal as $N$ increases. Intuitively, this occurs because the latent space controls the intended goal when the belief $b$ is roughly uniform, and then switches to controlling the preferred trajectory once the robot is confident. So if users change their goal after first convincing the robot, the latent space no longer contains actions that move towards this correct goal! We find that our proposed entropy loss function addresses this shortcoming: \textit{LA+SA+Entropy} users are able to input actions $z$ that alter the robot's goal. Our results support Sec.~\ref{sec:entropy}, and suggest that encouraging entropy at training time improves the robustness of the latent space.

\begin{figure*}[t]
	\begin{center}
		\includegraphics[width=1.8\columnwidth]{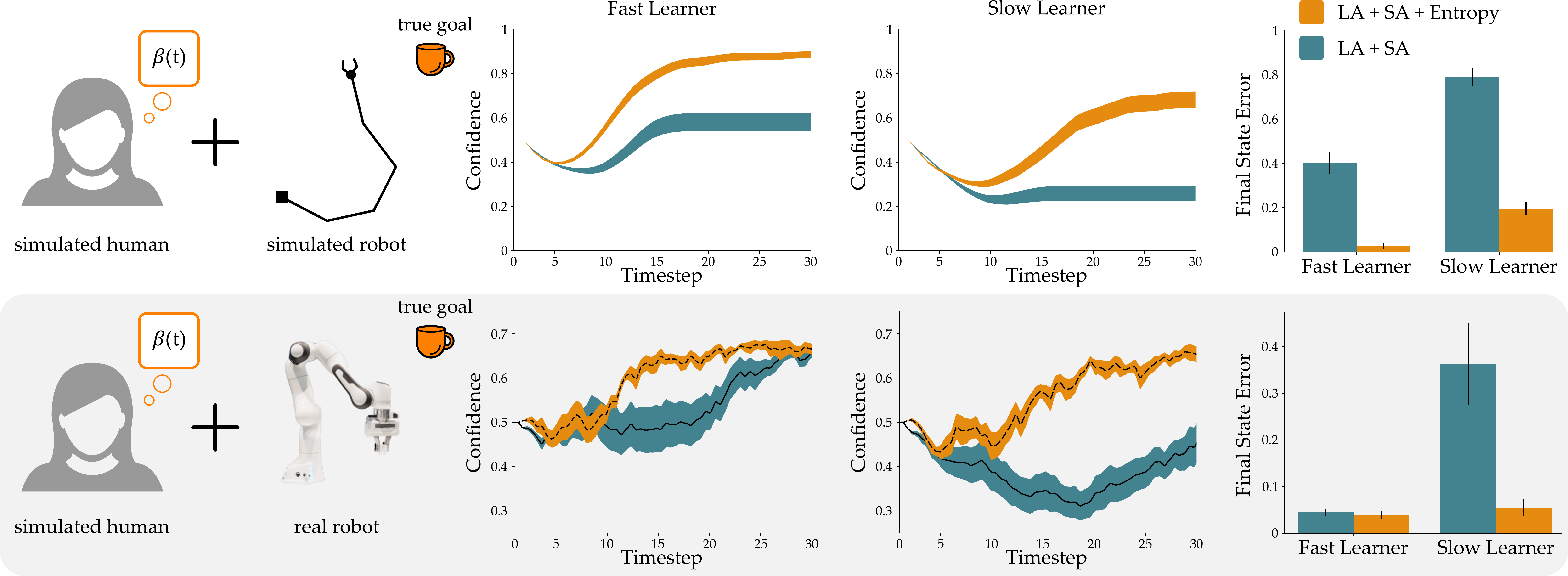}
		\caption{Simulated humans that learn how to teleoperate the robot. The human's rationality $\beta(t)$ is linear in time, and either increases with a high slope (\textit{Fast Learner}) or low slope (\textit{Slow Learner}). As the human learns, they get better at choosing inputs that best guide the robot towards their true goal. We find that latent actions learned with the entropy reward (\textit{LA+SA+Entropy}) are more versatile, so that the human can quickly undo mistakes made while learning.}
		\label{fig:sim3}
	\end{center}
	\vspace{-1.5em}
\end{figure*}

\subsection{Users that Learn within the Task}

We not only expect real users to change their mind when collaborating with the robot, but we also anticipate that these teleoperators will \textit{learn and improve} as they gain experience during the task. For instance, the user might learn that holding left on the joystick causes the robot to grasp the cup from the side, while holding right guides the robot towards a top grasp. To simulate this in-task learning, we set $\beta(t) = m\cdot t$, where the slope $m$ determines how quickly the user learns. All users start with random actions ($\beta = 0$), and either learn \textit{quickly} (high $m$) or \textit{slowly} (low $m$). We point out that slow learners may effectively ``change their mind'' multiple times, since they are unsure of how to control the robot.

Our findings are plotted in Fig.~\ref{fig:sim3}. Interestingly, we see that---for both fast and slow learners---\textit{LA+SA+Entropy} improves in-task performance. We attribute this improvement to the inherent \textit{versatility} of latent spaces that maximize entropy: as humans gain expertise, they can use these latent actions to quickly undo their mistakes and correct the robot's behavior.

\begin{figure}[t]
	\begin{center}
		\includegraphics[width=0.75\columnwidth]{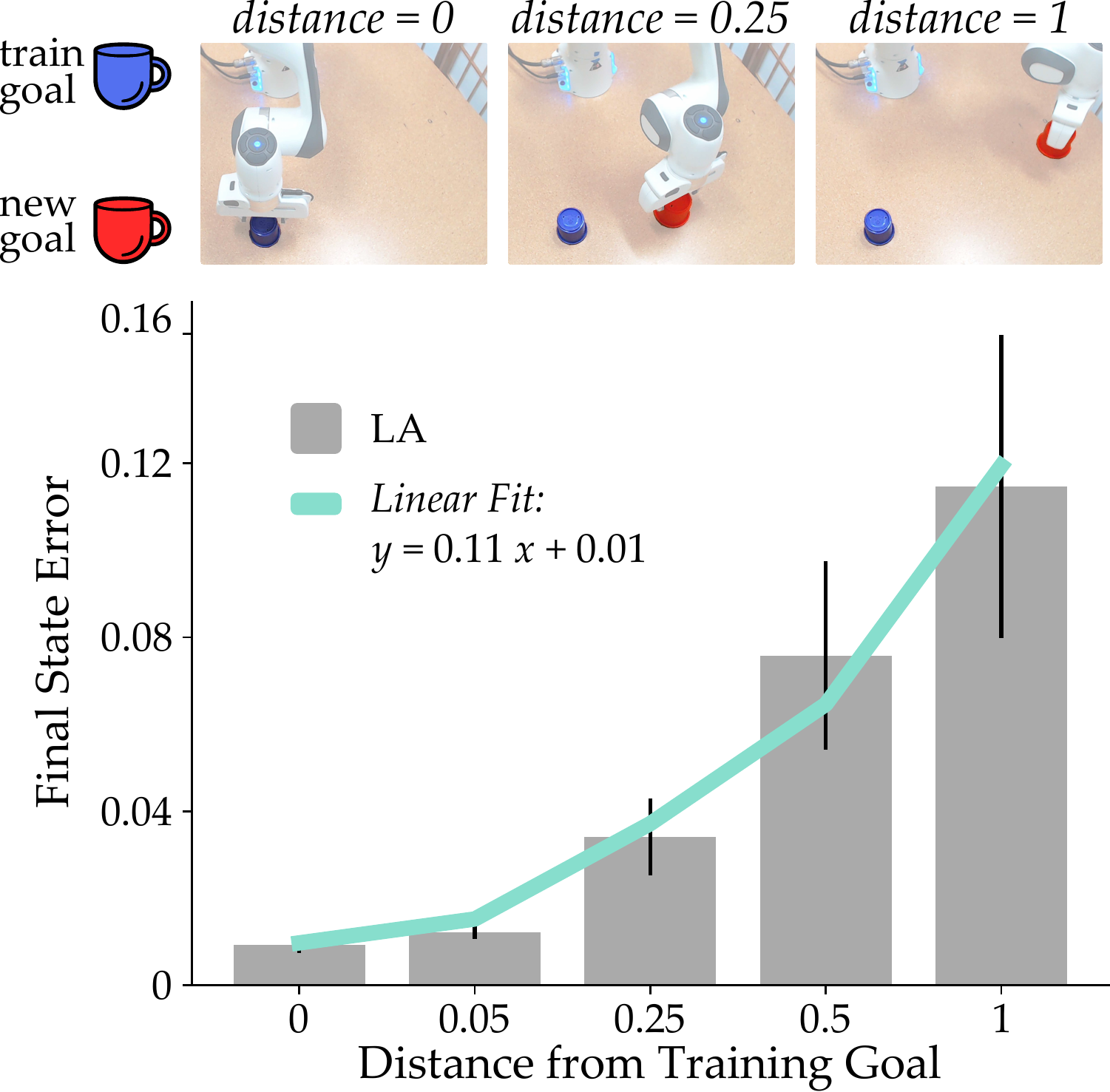}
	 	\vspace{-0.5em}
		\caption{Simulated humans teleoperating the real robot to reach new goals unseen at training time. Both axes are normalized by the distance between the training goal and the farthest new goal. The learned latent actions replicate demonstrations in the training dataset $\mathcal{D}$: we leverage known goals to \textit{extend} these latent actions into contexts where no training data is provided.}
		\label{fig:sim4}
	\end{center}
	\vspace{-2em}
\end{figure}

\subsection{Users Reaching for New Goals}

So far the simulated humans are grasping coffee cups that the robot observed at training time. Here we introduce \textit{new} coffee cups, and ask the users to reach for these goals \textit{without retraining} the latent space. If we make no changes to our approach, the latent actions can only ever reach the original cup---there are no demonstrations that grasp this new goal! We therefore geometrically map the robot's current context---which is outside of the decoder distribution---into a equivalent context defined relative to a known goal---where the robot can actually decode $z$. Our results on the real robot are summarized in Fig.~\ref{fig:sim4}. We observe a \textit{linear} relationship between distance and error ($y \approx 0.11x$, $R^2 = 0.98$): when the new coffee cup moves farther from the training cup, the robot's grasp becomes less accurate. But using goals as references has \textit{reduced this error}: without mapping the context to a known goal, the robot only ever reaches for the training goal ($y = x$).

\section{User Study}

Motivated by the application of assistive robotics, we designed a user study with \textit{eating tasks}. Participants teleoperated a $7$-DoF robotic arm with a $2$-DoF joystick to perform precise manipulation: users controlled the robot towards a goal plate, and then carefully adjusted the robot's motion to cut, stab, and scoop different foods (see Fig.~\ref{fig:front}).

\smallskip

\noindent\textbf{Experimental Setup.} Each participant attempted to complete two dishes: an \textit{Entree task} and a \textit{Dessert task}. In \textit{Entree}, users had to perform multiple precise motions at the same goal. Here participants (a) guided the robot towards a bowl with tofu, (b) cut off a slice of tofu, and (c) stabbed and scooped the slice onto their plate. In \textit{Dessert} the participants had to convey their preferences at multiple goals: they (a) stabbed a marshmallow in the middle goal, (b) scooped it through icing at the right goal, and then (c) dipped it in rice at the left goal before (d) setting the marshmallow on their plate. In both tasks subjects sat next to the robot, mimicking a wheelchair-mounted arm.

\smallskip
\noindent\textbf{Independent Variables.} We conducted a 2x2 factorial design that separately varied  \textit{Control Interface} and \textit{Robot Assistance}.

For the control interface, we tested a state-of-the-art direct teleoperation scheme (\textit{Retargetting}), where the user's joystick inputs map to the 6-DoF end-effector twist of the robot \cite{rakita2017motion}. We compared this direct teleoperation baseline to our learned \textit{Latent Actions}: here the robot interprets the meaning of the human's inputs based on the current context.

For robot assistance, we tested \textit{With} and \textit{Without Shared Autonomy}. We implemented the shared autonomy algorithm from \cite{javdani2018shared}, which assists the robot towards likely human goals.

Crossing these factors, we totaled 4 different conditions: 
\begin{itemize}
    \item Retargeting \textbf{(R)}
    \item Retargetting + Shared Autonomy \textbf{(R+SA)}
    \item Latent Actions \textbf{(LA)}
    \item Latent Actions + Shared Autonomy \textbf{(LA+SA)}
\end{itemize}
The \textit{LA+SA} condition is our proposed approach (Algorithm~\ref{alg:LA+SA}).

\smallskip

\noindent\textbf{Model Training.} We provided kinesthetic demonstrations $\mathcal{D}$ that guided the robot towards each plate, and then performed cutting, stabbing, and scooping motions at these goals. The robot learned the latent space (\textbf{LA}) from a total of $20$ minutes of kinesthetic demonstrations.

\smallskip

\begin{figure*}[t]
	\begin{center}
		\includegraphics[width=1.85\columnwidth]{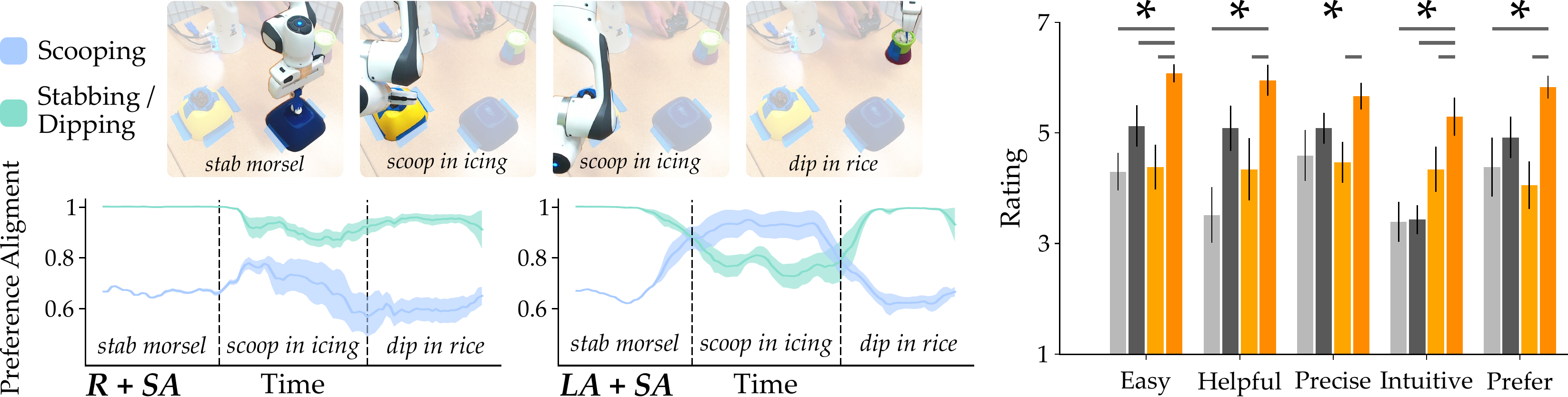}
		\caption{(Left) The \textit{Dessert} task consists of 3 phases: stabbing the morsel, scooping it in icing, and dipping it in rice. We identified the end-effector directions needed to complete these fine-grained preferences, and plotted the average dot product between the desired and actual directions (\textit{Preference Alignment}). With \textbf{R+SA}, users executed the entire task in a stabbing/dipping orientation. By contrast, with \textbf{LA+SA}, users correctly adjusted the scooping preference in the second phase of the task. (Right) We plot the results of our 7-point Likert scale surveys. Color to method mappings are consistent with Fig.~\ref{fig:error}.}
		\label{fig:dessert}
	\end{center}
	\vspace{-2em}
\end{figure*}

\noindent\textbf{Dependent Measures -- Objective.}
We recorded the amount of time users took to complete each task (\emph{Total Time}), as well as the amount of time spent without providing joystick inputs (\emph{Idle Time}). We also computed proxy measures of the high-level goal accuracy and low-level preference precision. For goals, we measured the robot's total distance to the closest plate throughout the task (\emph{Goal Error}). For preferences, we recorded the dot product between the robot's actual end-effector direction and the true end-effector directions needed to precisely cut, stab, and scoop (\textit{Preference Alignment}).

\smallskip

\noindent\textbf{Dependent Measures -- Subjective.} We administered a $7$-point Likert scale survey after each condition. Questions were organized along five scales: how \textit{Easy} it was to complete the tasks, how \textit{Helpful} the robot was, how \textit{Precise} their motions were, how \textit{Intuitive} the robot was to control, and whether they would use this condition again (\textit{Prefer}).

\smallskip

\noindent\textbf{Participants and Procedure.}
We recruited $10$ subjects from the Stanford University student body to participate in our study ($4$ female, average age $23.5 \pm 2.15$ years). All subjects provided informed written consent prior to the experiment. We used a within-subjects design: each participant completed both tasks with all four conditions (the order of the conditions was counterbalanced). Before every trial, users practiced teleoperating the robot with the current condition for up to $5$ minutes.

\smallskip

\noindent\textbf{Hypotheses.} We tested three main hypotheses:
\begin{displayquote}
\textbf{H1.} \emph{Users controlling the robot with Shared Autonomy (\textbf{SA}) will more accurately maintain their goals.}
\end{displayquote}
\begin{displayquote}
\textbf{H2.} \emph{Latent Actions (\textbf{LA}) will help users more precisely execute their preferences.}
\end{displayquote}
\begin{displayquote}
\textbf{H3.} \emph{Participants will complete the task most efficiently with combined \textbf{LA+SA}.}
\end{displayquote}

\smallskip

\begin{figure}[t]
	\begin{center}
		\includegraphics[width=1\columnwidth]{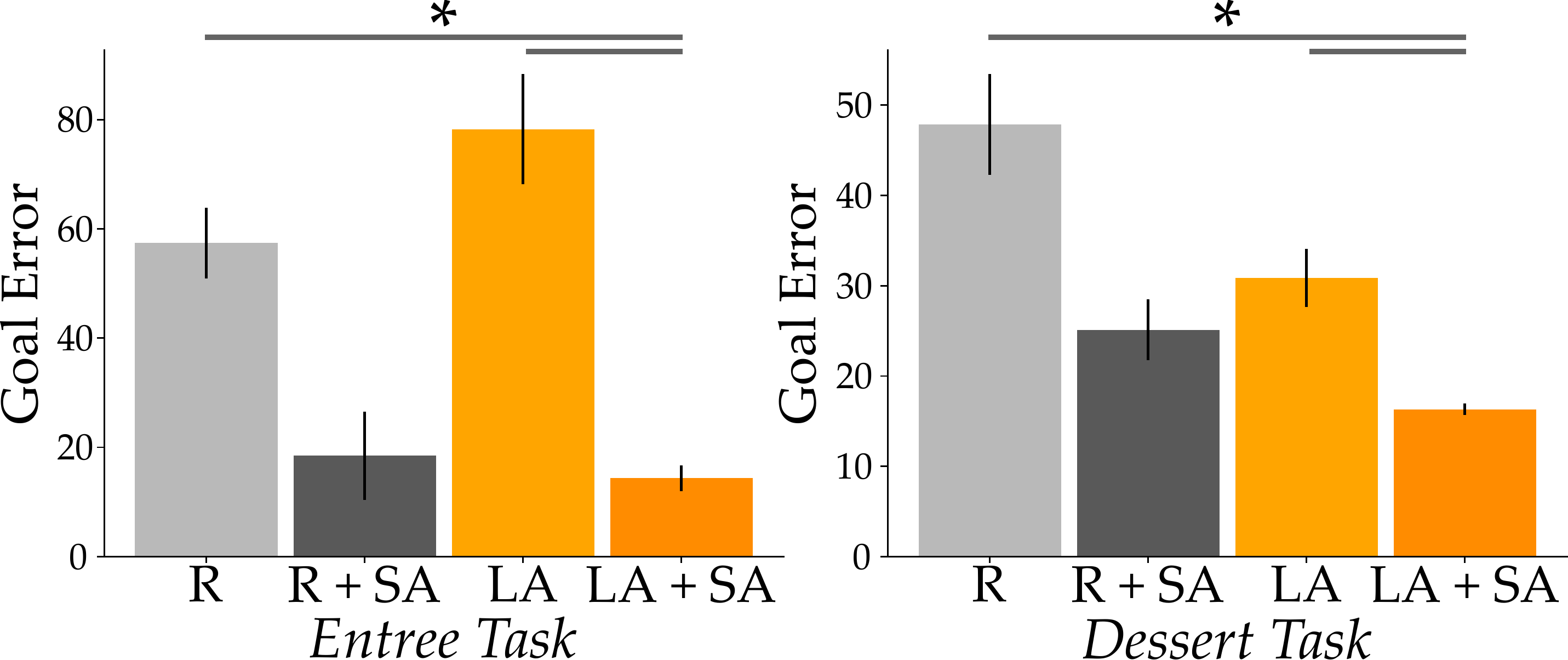}
		 	\vspace{-1.5em}
		\caption{Error between the nearest goal and the end-effector's current position. Adding \textbf{SA} decreased this error across both mapping strategies (\textbf{R} and \textbf{LA}).}
		\label{fig:error}
	\end{center}
 	\vspace{-1.em}
\end{figure}

\noindent\textbf{Results -- Objective.} To explore \textbf{H1}, we analyzed the \textit{Goal Error} for methods with and without \textbf{SA} (see Fig.~\ref{fig:error}). Across both tasks, users interacting with \textbf{SA} reached their intended goals \textit{significantly more accurately} $(F(1,18)= 29.9, p<.001)$. Breaking this down by condition, users incurred \emph{less} error with \textbf{LA+SA} than with \textbf{LA} ($p<.001$), and---similarly---users were more accurate with \textbf{R+SA} than with \textbf{R} ($p<.05$).

So shared autonomy helped users more accurately maintain their goals---but were participants able to complete the precise manipulation tasks at those goals? We visualize the \textit{Preference Alignment} for \textit{Dessert} in Fig.~\ref{fig:dessert}, specifically comparing \textbf{R+SA} to \textbf{LA+SA}. We notice that---when using direct teleoperation---participants remained in a stabbing preference throughout the task. By contrast, users with latent actions \textit{adjusted} between preferences: stabbing the marshmallow, scooping it in icing, and dipping it in rice. These results support \textbf{H2}, suggesting that \textbf{LA} enables users to express their preferences.

\begin{figure}[t]
	\begin{center}
		\includegraphics[width=1\columnwidth]{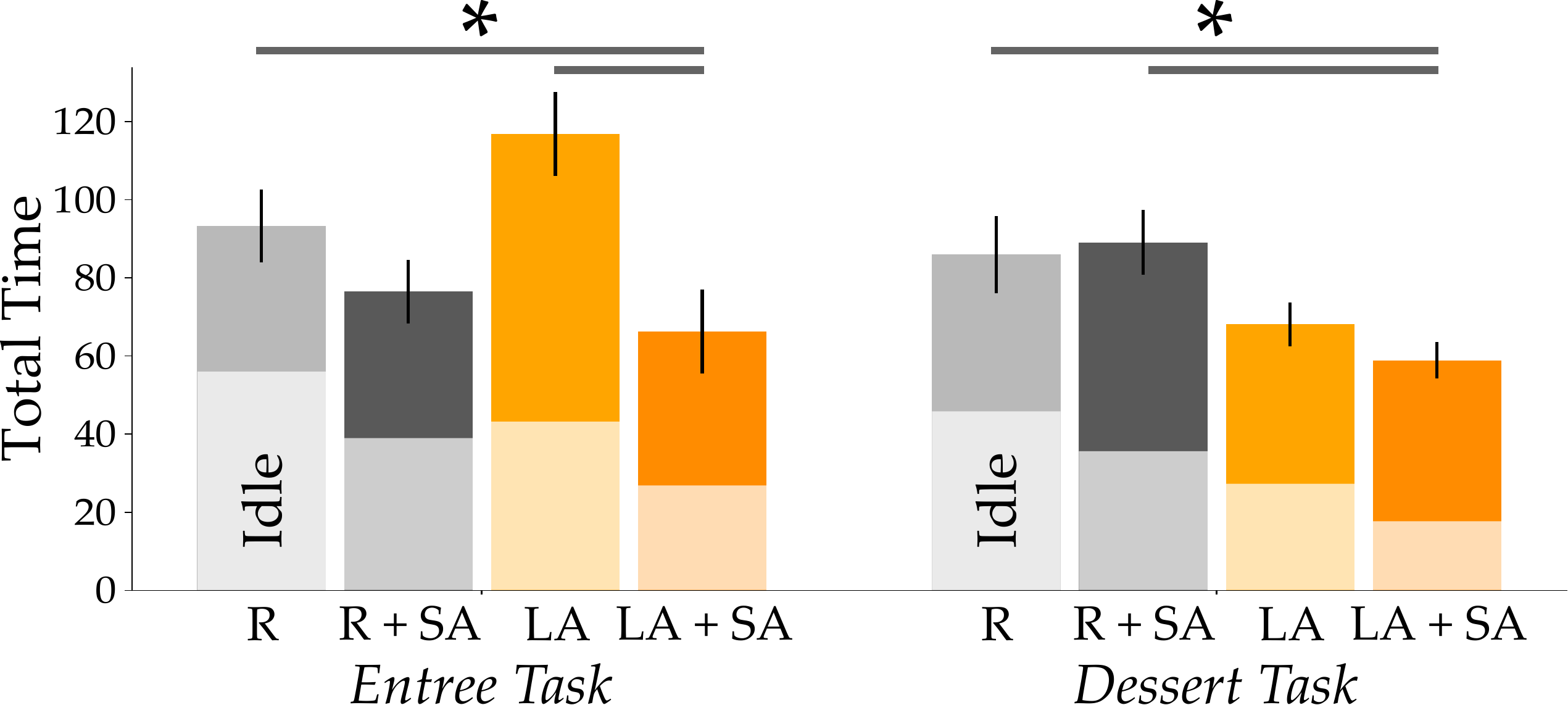}
	 	\vspace{-1.5em}
		\caption{Time taken to complete the task (solid) and time spent idle (light). Users completed both eating tasks most efficiently with \textbf{LA+SA}.}
		\label{fig:time}
	\end{center}
	\vspace{-2em}
\end{figure}

Now that we know the benefits of \textbf{SA} and \textbf{LA} individually, what happens when we focus on their combination? Inspecting Fig.~\ref{fig:time}, participants using \textbf{LA+SA} were able to complete both tasks more efficiently. Summing times across both tasks, and then performing pair-wise comparisons between each condition, we found that \textbf{LA+SA} outperformed the alternatives for both \textit{Total Time} ($p < .05$) and \textit{Idle Time} ($p < .05$).

\smallskip

\noindent\textbf{Results -- Subjective.} We find further support for \textbf{H3} in the user's feedback. The results of t-tests comparing \textbf{LA+SA} to the other conditions are reported in Fig.~\ref{fig:dessert} (where an $*$ denotes $p<.05$). Responses suggest that users were most ``comfortable" when performing precise manipulation with \textbf{LA+SA}.

\smallskip

\noindent\textbf{Limitations.} The $10$ participants in our user study were \textit{able-bodied}, and had equal practice time with each tested condition. By contrast, people applying these assistive control strategies on a daily basis will have significantly more experience and expertise, which could bias their skill towards conditions that are more familiar. We also recognize that our target population will likely provide joystick inputs with higher noise levels.

\section{Conclusion}

We focused on assistive teleoperation scenarios where the the human needs to control the robot's high-level goal and fine-grained motion. By combining shared autonomy with latent actions, we developed a method that constrains the human to their goal while embedding the robot's actions to precise sub-manifolds. Importantly, the meaning of the human's inputs changes as the robot becomes more confident---refining from coarse movements to careful adjustments. Our simulations and user studies suggest that this combined approach enables users to efficiently complete dexterous eating tasks.

\section*{Acknowledgments}
We thank Jeff Z. HaoChen for his early contributions to this work. We also acknowledge funding
by NSF grant \#1241349.

\bibliographystyle{plainnat}
\bibliography{references}

\end{document}